\documentclass[10pt,journal,compsoc]{IEEEtran}

\ifCLASSOPTIONcompsoc
  \usepackage[nocompress]{cite}
\else
  \usepackage{cite}
\fi
\ifCLASSINFOpdf
\else
\fi

\usepackage{xcolor}
\usepackage{amssymb}
\usepackage{adjustbox}
\usepackage{booktabs} 
\usepackage{amsmath}
\usepackage{caption}
\usepackage{multirow}
\usepackage{makecell}
\usepackage{setspace}

\usepackage{dsfont}

\usepackage{bm}
\newcommand{\PC}[1]{\ignorespaces}
\newcommand{\CP}[1]{\ignorespaces}

\newcommand{\ie}{\textit{i.e.}}
\newcommand{\eg}{\textit{e.g.}}
\newcommand{\Eg}{\textit{E.g.}}

\usepackage[linesnumbered,ruled,vlined]{algorithm2e}
\usepackage{algorithmicx}
\usepackage{algpseudocode}
\SetKwInput{KwInput}{Input}                
\SetKwInput{KwOutput}{Output}
\SetKwInput{KwParameters}{Learnable parameters}                
\SetKwProg{Init}{init}{}{}
\DeclareMathOperator*{\argmax}{arg\,max}

\DeclareSymbolFont{bbold}{U}{bbold}{m}{n}
\DeclareSymbolFontAlphabet{\mathbbold}{bbold}

\begin{document}
%
\title{Approaching Test Time Augmentation \\ in the Context of Uncertainty Calibration \\ for Deep Neural Networks}
%
%
%
%

\author{P. Conde,
        T. Barros,~\IEEEmembership{Student Member,~IEEE,}
        R.L. Lopes,
        C. Premebida,~\IEEEmembership{IEEE Member,}
and~U.J. Nunes,~\IEEEmembership{Senior,~IEEE}
\IEEEcompsocitemizethanks{\IEEEcompsocthanksitem P. Conde, T. Barros, C. Premebida and U.J. Nunes are with the Institute of Systems and Robotics, Department of Electrical and Computer Engineering, University of Coimbra.\protect\\
E-mail: \{pedro.conde,~tiagobarros,~cpremebida,~urbano\}@isr.uc.pt
\IEEEcompsocthanksitem R.L. Lopes is with Critical Software, S.A. \\ E-mail: rui.lopes@criticalsoftware.com }
}

%
%

\markboth{Submitted to \textit{IEEE} Transactions on Pattern Analysis and Machine Intelligence}%
{Conde \MakeLowercase{\textit{et al.}}: Approaching Test Time Augmentation in the Context of Uncertainty Calibration for Deep Neural Networks}
%



\IEEEtitleabstractindextext{%
\begin{abstract}
With the rise of Deep Neural Networks, machine learning systems are nowadays ubiquitous in a number of real-world applications, which bears the need for highly reliable models. This requires a thorough look not only at the accuracy of such systems, but also at their predictive uncertainty. Hence, we propose a novel technique (with two different variations, named \textit{M-ATTA} and \textit{V-ATTA}) based on test time augmentation, to improve the uncertainty calibration of deep \CP{learning-based}models for image \CP{classifiers} classification. By leveraging an adaptive weighting system, \textit{M/V-ATTA} improves uncertainty calibration without affecting the model’s
accuracy. 
The performance of these techniques is evaluated by considering diverse metrics related to uncertainty calibration, demonstrating their robustness.
Empirical results, obtained on CIFAR-10, CIFAR-100, Aerial Image Dataset, as well as in two different scenarios under \textit{distribution-shift}, indicate that the proposed methods outperform several state-of-the-art \textit{post-hoc} calibration techniques. Furthermore, the methods proposed also show improvements in terms of predictive entropy on \textit{out-of-distribution} samples. Code for \textit{M/V-ATTA} available at: \text{https://github.com/pedrormconde/MV-ATTA}
\end{abstract}

\begin{IEEEkeywords}
Uncertainty Calibration, Reliability, Probabilistic Interpretation, Test Time Augmentation, Deep Neural Networks.
\end{IEEEkeywords}}


\maketitle

\IEEEdisplaynontitleabstractindextext

%
\IEEEpeerreviewmaketitle

\ifCLASSOPTIONcompsoc
\IEEEraisesectionheading{\section{Introduction}\label{sec:introduction}}
\else
\section{Introduction}
\label{sec:introduction}
\fi

%
%
%
%
\IEEEPARstart{D}{eep} Neural Networks (DNNs) changed the paradigm with regards to the applicability of machine learning (ML) systems to real-world scenarios. Consequently, \CP{these} deep learning (DL) models are now present in critical application domains (\eg, \CP{transportation}autonomous driving, medicine, remote sensing, robotics), where bad decision-making can bear potentially drastic consequences. This requires \CP{that} DNNs to be \CP{are} not only highly accurate, but also highly \textit{reliable} - \CP{the user must}decision-makers should be able to ``trust" the predictions of these models. This leads us to the problem of \textit{uncertainty calibration} (also referred as \textit{confidence calibration} \cite{kuppers2020multivariate} or simply \textit{calibration} \cite{guo2017calibration}): ensuring that the confidence output generated by the DL model - that translates as the confidence the model has in the prediction that is making - realistically represents the true likelihood of correctness. \PC{The problem of uncertainty calibration will be properly formulated in Section \ref{Background};} For the sake of intuition, a calibrated model would for example, in the long run, correctly classify $70\%$ of those predictions that have a confidence value of $0.7$ associated (the formalization of the uncertainty calibration problem will be described in Section \ref{Background}). \PC{a proper calibrated model having a confidence value of, for example, $0.7$ would, in the long run, correctly classify $70\%$ of its predictions.} This accurate quantification of predictive uncertainty results in reliable confidence values associated with each prediction, and therefore, in \CP{overall} a more reliable model. As such, it is important to understand how well calibrated \CP{are} modern DNNs are in representative scenarios.\\
\indent \PC{Nevertheless, it is reasonable to ask \CP{But} how well calibrated \CP{are} modern DNNs are.} Although increasingly accurate, modern DL architectures have been found to be tendentiously \textit{uncalibrated} \cite{guo2017calibration, ovadia2019can}. Furthermore, ``modern neural networks exhibit a strange phenomenon:
probabilistic error and miscalibration worsen even as classification error is reduced" \cite{guo2017calibration}. For this reason, the goal of this work is to improve the uncertainty calibration of DNNs in the task of image classification, by proposing a novel accuracy-consistent weighted test time augmentation method.\\
\indent  Test time augmentation is a general methodology that leverages data augmentation techniques, to create multiple samples from the original input, at inference (\ie, at test time). Therefore, test time augmentation methods can be applied to previously (already) trained models. \CP{, since, in this case, the augmentation process is not applied during the training phase.} The methods introduced in this work combine the use of test time augmentation with a custom weighting system, guaranteeing that the accuracy of the original DL model is not corrupted, while still being optimized to improve uncertainty calibration. This builds, partially, on the work done in \cite{Conde_2022_BMVC}, proposing both a reformulated version - \textit{V-ATTA} (Vector Adaptive Test Time Augmentation) - of the preliminary method presented in \cite{Conde_2022_BMVC} and also a new generalized version - \textit{M-ATTA} (Matrix Adaptive Test Time Augmentation) - with a broader and extended empirical evaluation. \\
\indent \textit{M/V-ATTA} is evaluated by leveraging the benchmark CIFAR-10/CIFAR-100 \cite{krizhevsky2009learning} datasets, as well as the benchmark Aerial Image Dataset (AID) \cite{xia2017aid}, to create different representative experimental setups. The results are compared with \CP{a} state-of-the-art \textit{post-hoc} calibration methods, with respect to different evaluation metrics, related to both \textit{common} and \textit{strong} uncertainty calibration (see Section \ref{Background} for further details), and also \textit{out-of-distribution} (OOD) predictive entropy. \\
\indent With this work, we address a gap in the existing literature, concerning \textit{data-centric post-hoc} methodologies for uncertainty calibration \textit{i.e.}, \textit{post-hoc} methods that leverage the manipulation of data (such as augmentations) to obtain better calibrated predictions. This distinction is important, \CP{, since}because most \textit{post-hoc} calibration methods focus solely on altering the predictive behaviour through the manipulation of individual prediction/logits vectors \cite{guo2017calibration, zadrozny2002transforming, zadrozny2001obtaining, kull2019beyond}. \\ 
\indent \textbf{Contribution}: 
We propose a novel calibration technique - with two different variations (\textit{M/V-ATTA}) - based on test time augmentation, that preserves the accuracy of the deep models with which is applied, while being shown to improve uncertainty calibration, outperforming state-of-the-art \textit{post-hoc} calibration methods in most cases. Furthermore, \CP{Our}the methods presented here can be used with previously trained DNNs, which is advantageous in terms of applicability. 

    \PC{\item A brief study of the effects that different policies of augmentation have in the performance of \textit{M/V-ATTA}, in the context of uncertainty calibration.}

\section{Related Work}

The authors in \cite{guo2017calibration} introduce the topic of uncertainty calibration to the DL community, by evaluating how well calibrated are different modern DNNs. \CP{, using different datasets (from both computer vision and natural language processing applications).} As previously stated, the authors argue that, although more accurate, modern DNNs exhibit problems of miscalibration, often more severe than those found in older - and less accurate - architectures. Several \textit{post-hoc} calibration techniques - particularly designed to address uncertainty calibration without the need of re-training the original DL models - like \textit{temperature scaling} (an extension of the \textit{Platt scaling} algorithm \cite{platt1999probabilistic, niculescu2005predicting}), \textit{histogram binning} \cite{zadrozny2001obtaining} and \textit{isotonic regression} \cite{zadrozny2002transforming}, are described to address this problem. 
\\
\indent Other approaches, like  approximate Bayesian models \cite{graves2011practical, gal2016dropout} and some regularization techniques \cite{pereyra2017regularizing, muller2019does}, have also been used in the context of uncertainty calibration \cite{ovadia2019can}. Nonetheless, these approaches require building new more complex models or modifying and re-training pre-existing ones, contrarily to the previously mentioned \textit{post-hoc} calibration methods and the proposed \textit{M/V-ATTA}. \\
\indent For the evaluation of uncertainty calibration, one of the most popular metrics is the ECE \cite{naeini2015obtaining}, where the bin-wise difference between the accuracy and the confidence outputs of a given model is evaluated. Although intuitive, some limitations of this metric have been identified in relevant literature \cite{nixon2019measuring, gupta2021top, widmann2019calibration}, with regard to, for example, its intractability and dependence on the chosen binning scheme. Furthermore, because ECE is not a proper scoring rule \cite{gneiting2007strictly}, there can be found trivial uninformative solutions that obtain an optimal score, like always returning the marginal probability (see example in Supplementary Material - ``Notes on the ECE"). A popular alternative to evaluate uncertainty calibration is the Brier score \cite{brier1950verification}; because it is a proper scoring rule and  overcomes some of the identified limitation of the ECE, this metric has been increasingly used by the scientific community \cite{ovadia2019can, tian2021geometric, moon2020confidence}.\\
\indent Test time augmentation methods have been gaining some attention in the last few years, for example in biomedical applications \cite{wang2019aleatoric, wang2018test, moshkov2020test}. Nonetheless, we see that some potential in this type of technique is still under-researched and, to the best of our knowledge, all relevant literature fails to address its effect on calibration-specific metrics like the Brier score or the ECE (with exception of our preliminary work done in \cite{Conde_2022_BMVC}). Contrarily to other recent approaches to test time augmentation \cite{lyzhov2020greedy, kim2020learning,shanmugam2021better}, the method proposed in this work does not focuses on improving the accuracy of DNNs, but instead on improving their uncertainty calibration without altering the original accuracy. In fact, the authors in \cite{shanmugam2021better} show that some forms of test time augmentation may produce corrupted predictions - which can ultimately worsen the model's performance - reinforcing the need for a test time augmentation-based methodology that does not alter the predicted class, while still calibrating its confidence value.


\section{Background}
\label{Background}

\textbf{Notation}: we will use bold notation to denote vectors, like $\mathbf{p}=~(p_1,\ldots,p_{k})$; the $i$-nth element of some vector $\mathbf{p}$ will be referred as $\mathbf{p}_{\{i\}}:=p_i$; the $\downarrow$ symbol, associated with a given metric, informs that a lower value of such metric represents a better performance; the $\odot$ symbol represents the Hadamard product; $\sigma: \mathbb{R}^k \rightarrow \Delta_k$ represents the \textit{softmax} function. These remarks are valid for all sections of the article; other remarks on notation will be given along the text, when found relevant.  \\
\\
\PC{\indent In this section we present common metrics for assessing uncertainty calibration, as well as some \textit{post-hoc} calibration methods - \textit{temperature scaling}, \textit{isotonic regression} and \textit{histogram binning} - that will be used as baseline for comparison with the approaches proposed in this work.\\}
\indent In this section we discuss the problem of uncertainty calibration and present some evaluation metrics commonly used in this context. \PC{\\
\indent We will start by introducing to the concept of uncertainty calibration.} In relevant literature, the concept of uncertainty calibration related to DL systems, in a multi-class scenario, is often defined in two different ways. The most common way is the one presented in \cite{guo2017calibration}, where the multi-class scenario is considered as an extension of the binary problem in a \textit{one vs all} approach, taking in account only the calibration of the highest confidence value for each prediction. Nonetheless, some works consider the more general definition present in \cite{zadrozny2001obtaining}, that takes into account all the confidence values of the predicted probability vector. Like in \cite{widmann2019calibration}, we make the respective distinction between a \textit{calibrated} model and a \textit{strongly calibrated} model, in the following definitions. Notice that in the case of a binary classifier such definitions are equivalent.\\
\indent Let us consider a pair of random variables $(X,Y)$, where $X$ represents an input space (or feature space) and $Y$ the corresponding set of true labels. Let us now take a model $f: X \rightarrow \Delta_k$, with $\Delta_k=~\{(p_1,\ldots,p_{k}) \in [0,1]^k : \sum_{i=1}^{k} p_i =1 \}$ being a probability simplex (this setting corresponds to a classification problem with $k$ \CP{different} classes). The model $f$ is considered \textbf{calibrated} if

\begin{align}
\label{calibration}
    \mathds{P} [Y = \argmax_{i \in \{1,\ldots,k\}} f(X)
    \ | \   \max_{i \in \{1,\ldots,k\}} f(X)] =
    \max_{i \in \{1,\ldots,k\}} f(X).
\end{align}
Additionally, the model $f$ is considered \textbf{strongly calibrated} if

\begin{align}
\label{strong_calibration}
    \mathds{P} [Y = y \ | \  f(X)_{\{y\}}] = f(X)_{\{y\}}.
\end{align}

\indent As stated in \cite{guo2017calibration}, achieving perfect calibration is impossible in practical settings. Furthermore, the probability values in the left hand side of both \eqref{calibration} and \eqref{strong_calibration} cannot be computed using finitely many samples, which motivates the need for evaluation metrics to assess uncertainty calibration.

\subsection{Brier score $\downarrow$}

Brier score \cite{brier1950verification} is a proper scoring rule  \cite{gneiting2007strictly} that computes the squared error between a predicted probability and its true response, hence its utility to evaluate uncertainty calibration. 
For a set of $N$ predictions we define the \textbf{Brier score} as
\begin{align}
\label{brier}
    \text{BS} = \frac{1}{N} \sum_{j=1}^N (p^j - o^j)^2,
\end{align}
where $p^j$ is the highest confidence value of the prediction $j$ and $o^j$ equals 1 if the true class corresponds to the prediction and 0 otherwise.

The previous definition is useful to assess calibration (in the more common sense). Nonetheless, \cite{brier1950verification} also presents a definition for Brier score in multi-class scenario that is suitable to assess strong calibration.
For a problem with $k$ different classes and a set of $N$ predictions we define the \textbf{multi-class Brier score} (mc-Brier score) as
\begin{align}
\label{multi_brier}
    \text{mc-BS}= \frac{1}{N} \sum_{j=1}^N \sum_{i=1}^k (p_i^j - o_i^j)^2,
\end{align}
where $p_i^j$ is the confidence value in the position $i$ of the $j$-nth prediction probability vector and $o_i^j$ equals 1 if the true label equals $i$ and 0 otherwise.\\
\indent We refer to \cite{murphy1973new} and \cite{blattenberger1985separating} for some thorough insights about the interpretability and decomposition of the Brier score. 

\subsection{Expected Calibration Error $\downarrow$}

To compute the ECE \cite{naeini2015obtaining} we start by dividing the interval $[0,1]$ in $M$ equally spaced intervals. Then a set of bins $\{ B_1, B_2, \ldots, B_M \}$ is created, by assigning each predicted probability value to the respective interval. The idea behind this measurement is to compute a weighted average of the absolute difference between accuracy and confidence in each bin $B_i$ ($i=1,\ldots,M$). 
We define the \textbf{confidence per bin} as 
\begin{align}
\label{confidence}
    conf(B_i) = \frac{1}{|B_i|} \sum_{j \in B_i} p^j,
\end{align}
where $p^j$ is the highest confidence value of the prediction $j$. 
The \textbf{accuracy per bin} is defined as
\begin{align}
\label{accuracy}
    acc(B_i) = \frac{1}{|B_i|} \sum_{j \in B_i} o^j,
\end{align}
where $o^j$ equals 1 if the true class corresponds to the prediction and 0 otherwise.
For a total of $N$ predictions and a binning scheme $\{ B_1, B_2, \ldots, B_M \}$, the \textbf{ECE} is defined as
\begin{align}
\label{ece}
   \text{ECE} = \sum_{i=1}^M \frac{|B_i|}{N} | conf(B_i) - acc(B_i) |.
\end{align}

\goodbreak
\section{Proposed Methodology}
\label{section:method}

\begin{figure*}[]
    \centering
    \includegraphics[width=1.08\textwidth]{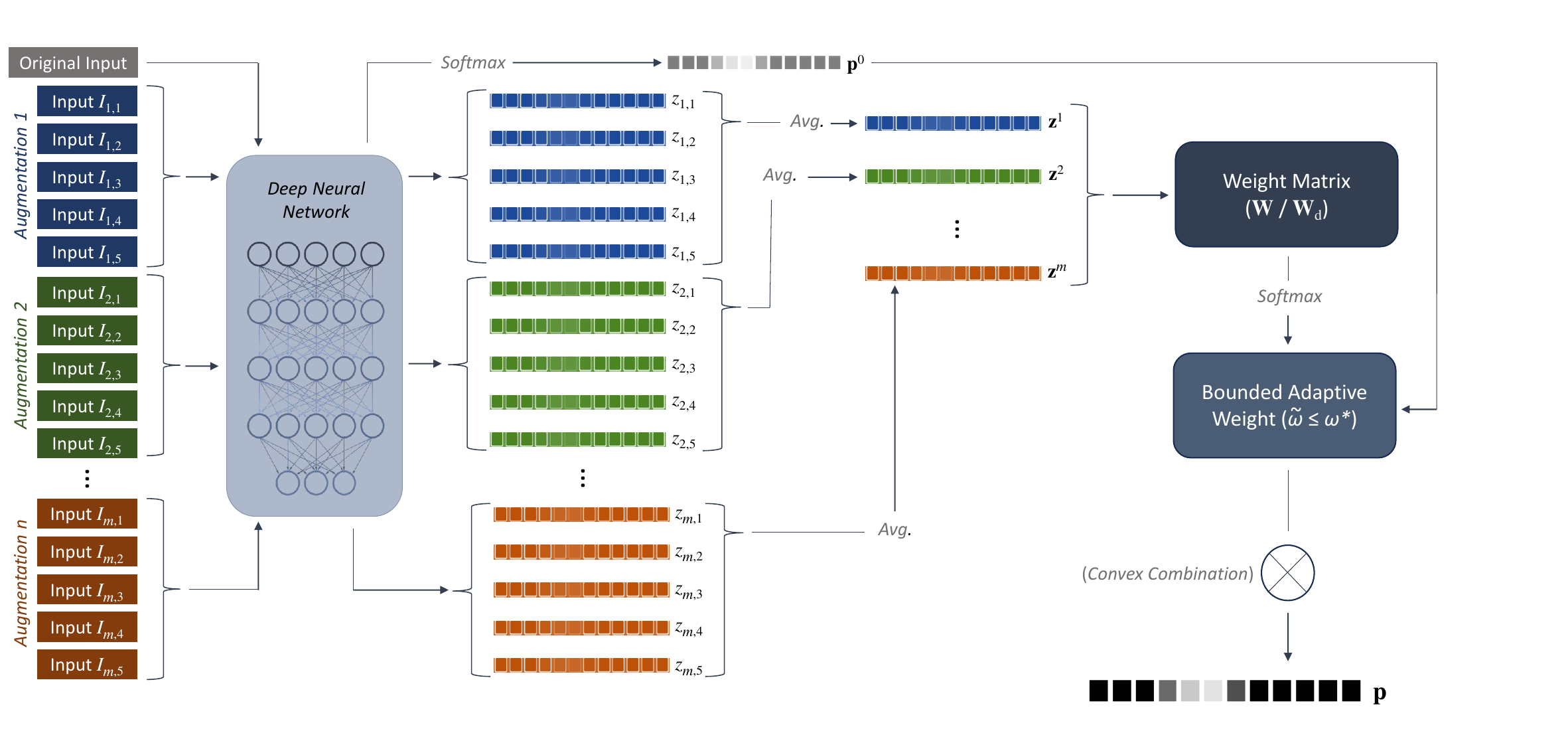}  
    \caption{Overview of the common structure of both \textit{M-ATTA} and \textit{V-ATTA} - from a ``high-level" perspective - to serve as a general graphical support for the detailed description presented in Subsections \ref{subsect:M-Atta} and \ref{subsection:V-Atta}. In this figure, we consider (for the purpose of illustration) $n_i = 5, \forall i \in \{1,2,m\}$, although this value can take the form of any natural number, as clarified by the detailed description. }
    \label{fig_method}
\end{figure*}

In this section we introduce \textit{M-ATTA} and \textit{V-ATTA}. Both methods leverage the use of test time augmentation, combined with a custom adaptive weighting system. \textit{V-ATTA} can be interpreted as a restricted version of the more general \textit{M-ATTA}.\\

\subsection{Motivation and key insights}

\indent Test time augmentation leverages the manipulation of data at inference time, augmenting it through different transformations, therefore inducing variability in that data, resulting in multiple heterogeneous predictions. Intuitively, intelligently combining different predictions - that result from ``different perspectives" of the same data sample - is expected to calibrate the way a model estimates how confident/uncertain is, when predicting the class of that data sample. Test time augmentation can thus be leveraged in various forms of - what we can define as - \textit{data-centric ensemble} techniques, which aggregate predictions derived from different data inputs. This is exemplified by the proposed M/V-ATTA. Contrarily to \textit{model-centric ensembles} (that ensemble predictions resulting from different models), like \textit{deep ensembles} \cite{lakshminarayanan2017simple, rahaman2021uncertainty}, the proposed approaches do not require the training of several DNNs for the same task, being therefore applicable to any previously trained DNN\CP{, as has been already observed}. \\
\indent \textit{M/V-ATTA} defines a type of \textit{data-centric ensemble} that is based on two general principles:
\begin{enumerate}
    \item The way each augmented prediction impacts the final prediction is defined trough a set of weights, that are derived based on a validation set. For \textit{V-ATTA}, only one weight per type of augmentation is optimized, whereas for \textit{M-ATTA}, $k$ weights per type of augmentation are optimized, one for each of the $k$ classes.\\
    \item In each prediction, an adaptive weight (whose upper bound is optimized in the aforementioned validation set) adjusts its value, controlling how the original prediction (the one that results from the non-augmented data sample) weights in the final prediction, ensuring that the \textit{maximum value of the final prediction} and the \textit{maximum value of the original prediction} represent the same class. In this way, the proposed methods work similarly to \textit{intra order-preserving functions} \cite{rahimi2020intra} (like is the case of \textit{temperature scaling}), although only preserving the order of the maximum, and not the overall intra order of the original prediction vector. Consequently, applying \textit{M/V-ATTA} will not have any effect on the accuracy of the DNN (see Supplementary Material - ``\textit{M/V-ATTA} accuracy consistency" - for the formal proof).
\end{enumerate}

Figure \ref{fig_method} provides a general illustration of the common structure of both \textit{M-ATTA} and \textit{V-ATTA}, to give a broad overview of the proposed techniques that are \CP{what is} - in further detail - described in Subsections \ref{subsect:M-Atta} and \ref{subsection:V-Atta}.

\subsection{M-ATTA}
\label{subsect:M-Atta}

\begin{algorithm}[]
 
    \KwInput{Augmented logits matrix ($\mathbf{Z} \in \mathbb{R}^{k , m}$), Original prediction ($\mathbf{p}^0 \in \Delta_k$)}
 
	\KwOut{Calibrated prediction ($\mathbf{p} \in \Delta_k$)}
    \KwParameters{$\mathbf{W}\in \mathbb{R}^{k , m}$, $\omega^* \in \mathbb{R}$}
  
    $\epsilon \gets 0.01$  
     
     $\tilde{\omega} \gets \omega^*$  
     
     $c_0 \gets \argmax_{i \in \{1,\ldots,k\}} \mathbf{p}^0$   

     
    \While{$c \neq c_0\  \land \ \tilde{\omega} >0$}{

         $\mathbf{p} \leftarrow (1-\tilde{\omega})\mathbf{p}^0 + \tilde{\omega} \text{ } \sigma(\mathbf{W} \odot \mathbf{Z}) I_m$  
         
         $c \gets \argmax_{i \in \{1,\ldots,k\}} \mathbf{p}$ 
         
         $\tilde{\omega} \gets \tilde{\omega} - \epsilon$  
     }
   \caption{M-ATTA (adaptive weight approximation)} \label{alg:m-atta}
 \end{algorithm}

Let us start by considering $m \in \mathbb{N}$ different types of augmentations. Because it is common that some augmentations have random parameters, it can be desirable to apply the same type of augmentation more than once; for this reason, let us consider as $n_i$ ($i=1,\ldots,m$) the number of times the $i$-nth type of augmentation is applied. As such, we will define for each original input $I_0$, the $j$-nth ($j=1,\ldots,n_i$) augmented input with the $i$-nth augmentation type, as $I_{i,j}$. \\
\indent We can now take into account the model $f: X \rightarrow \Delta_k$ (where $X$ is the input space and $k$ the number of classes) and consider $g: X \rightarrow \mathbb{R}^k$ as such that $f := g \circ \sigma$. \CP{With this}Consequently, we now define
\begin{align}
    \textbf{p}^0 = f(I_0), \qquad z_{i,j} = g(I_{i,j}),
\end{align}
\ie, $p^0$ is the probability vector associated with the original input and $z_{i,j}$ is the logit associated with the $j$-nth augmentation of the $i$-nth type.
Subsequently, we can define $\forall i \in [1,\ldots,m]$, 
\begin{align}
    \mathbf{z}^i = \frac{\sum_{j=1}^{n_i} \mathbf{z}_{i,j}}{n_i}
    \equiv \left(z^i_1, z^i_2, \ldots, z^i_k\right)
    \in \mathbb{R}^k,
\end{align}
and then construct the matrix
\begin{align}
    \mathbf{Z}= \left[ \mathbf{z}^i \right]_{i=1,\ldots,m}^\text{T} =
\left[ \begin{array}{ccccc}
z^1_1 & z^2_1 &  \cdots  & z^m_1 \medskip \\ 
z^1_2 & z^2_2   &  \cdots  & z^m_2   \\
  \vdots         &      \vdots       &     \ddots    & \vdots   \\
  z^1_k         &     z^2_k       &     \cdots    & z^m_k
\end{array}\right] \in \mathbb{R}^{k,m}.
\end{align}
Now, for some parameters 
\begin{align}
\omega^* \in  [0,1], \quad \mathbf{W}=
\left[ \begin{array}{ccccc}
\omega^1_1 & \omega^2_1 &  \cdots  & \omega^m_1 \medskip \\ 
\omega^1_2 & \omega^2_2   &  \cdots  & \omega^m_2   \\
  \vdots         &      \vdots       &     \ddots    & \vdots   \\
  \omega^1_k         &     \omega^2_k       &     \cdots    & \omega^m_k
\end{array}\right] \in \mathbb{R}^{k,m},
\end{align}
we finally define, for each prediction, the final prediction probability vector as
\begin{align}
    \mathbf{p}\left(\tilde{\omega}\right) = 
    (1-\tilde{\omega})\mathbf{p}^0 + \tilde{\omega} \text{ } \sigma(\mathbf{W} \odot \mathbf{Z}) I_m,
\end{align}
with
\begin{align}
\label{omega}
    \tilde{\omega} = \max \Big{\{} \omega \in [0,\omega^*] : \argmax_{i \in \{1,\ldots,k\}} \mathbf{p}\left(\tilde{\omega}\right) = \argmax_{i \in \{1,\ldots,k\}} \mathbf{p}^0 \Big{\}}.
\end{align}
$I_m \in \mathbb{R}^m$ represents an $m$-dimensional vector where every element equals 1 (remember also that $\sigma: \mathbb{R}^k \rightarrow \Delta_k$ represents the \textit{softmax} function). Additionally, the learnable parameters $\mathbf{W}\in \mathbb{R}^{k , m}$  and $\omega^* \in \mathbb{R}$ work, respectively, as an weight matrix and an upper bound for $\tilde{\omega}$. \\
\indent The value of $\tilde{\omega}$ may vary in each prediction, adapting in a way that prevents corruptions in terms of accuracy, according to the definition in \eqref{omega}. Both $\omega^* \in [0,1]$ and $\mathbf{W} \in \mathbb{R}^{k,m}$ can be optimized \CP{with}on a given validation set. In a practical scenario, the value $\tilde{\omega}$ is approximated as outlined in the algorithmic description presented in Algorithm \ref{alg:m-atta}. In our case $\epsilon$ in Algorithm \ref{alg:m-atta} is \CP{defined as}set to 0.01.

\subsection{V-ATTA}
\label{subsection:V-Atta}

\begin{algorithm}[]

    \KwInput{Augmented logits matrix ($\mathbf{Z} \in \mathbb{R}^{k , m}$), Original prediction ($\mathbf{p}^0 \in \Delta_k$)}
 
	\KwOut{Calibrated prediction ($\mathbf{p} \in \Delta_k$)}
    \KwParameters{$\mathbf{W}_d \in \mathbb{R}^{m , m}$, $\omega^* \in \mathbb{R}$}
  
    $\epsilon \gets 0.01$  
     
     $\tilde{\omega} \gets \omega^*$  
     
     $c_0 \gets \argmax_{i \in \{1,\ldots,k\}} \mathbf{p}^0$   

     
    \While{$c \neq c_0\  \land \ \tilde{\omega} >0$}{

         $\mathbf{p} \leftarrow (1-\tilde{\omega})\mathbf{p}^0 + \tilde{\omega} \text{ } \sigma (\mathbf{W}_d  \mathbf{Z}^{\text{T}}) I_m$  
         
         $c \gets \argmax_{i \in \{1,\ldots,k\}} \mathbf{p}$ 
         
         $\tilde{\omega} \gets \tilde{\omega} - \epsilon$  
     }
   \caption{V-ATTA (adaptive weight approximation)} \label{alg:v-atta}
 \end{algorithm}

With \textit{V-ATTA} we restrict the matrix $\mathbf{W}$ to a diagonal matrix
\begin{align}
    \mathbf{W}_d=
\left[ \begin{array}{ccccc}
\omega^1 & 0 &  \cdots  & 0 \medskip \\ 
0 & \omega^2   &  \cdots  & 0  \\
  \vdots         &      \vdots       &     \ddots    & \vdots   \\
  0        &     0       &     \cdots    & \omega^m
\end{array}\right] \in \mathbb{R}^{m,m},
\end{align}
and define the new prediction probability vector as
\begin{align}
    \mathbf{p}\left(\tilde{\omega}\right) = 
    (1-\tilde{\omega})\mathbf{p}^0 + \tilde{\omega} \text{ } \sigma (\mathbf{W}_d  \mathbf{Z}^{\text{T}}) I_m,
\end{align}
with $\tilde{\omega}$ still defined as in \eqref{omega}.\\
\indent In this case, the algorithmic description for the approximation of $\tilde{\omega}$ is represented in Algorithm \ref{alg:v-atta}. \CP{Once again, $\epsilon$ is 0.01.}

\section{Experiments and Results}
\label{adapttta:results}

\begin{table}[b]
    \centering
    \begin{tabular}{l||l}
    \toprule
        Aug. 1: \  $F + 5Cr$  &  Aug. 5: \ $F + 5Cr + 5 Ct $   \\
        Aug. 2: \  $F + 5B$ &  Aug. 6: \ $F + 5B + 5Ct$  \\
        Aug. 3: \ $5Cr + 5B$  & Aug. 7: \ $5Cr + 5B + 5 Ct$ \\
        Aug. 4: \ $F + 5Cr + 5B$ & Aug. 8: \ $F+5Cr+5B+5Ct$ \\
        \bottomrule
    \end{tabular}
    \caption{Composition of eight different augmentation policies used in this work. $F$ refers to a \textit{flip} transformation, $Cr$ to \textit{crop}, $B$ to \textit{brightness} and $Ct$ to \textit{contrast}. \Eg, Augmentation Policy 1 (Aug. 1) is composed by one \textit{flip} transformation and five \textit{crop} transformations.}
    \label{aug_policies}
\end{table}

\indent In all experiments, a ResNet-50 \cite{he2016deep} architecture was used. The achieved classification accuracy values were $94.21\%$, $72.78\%$ and $93.40\%$ \CP{, for}on CIFAR-10, CIFAR-100 and AID  test sets, respectively. The size of the training, validation and test sets are respectively: 50000, 1000, 9000, for the CIFAR-10/100 datasets, 7000, 1000, 2000, for the AID dataset. Uniform size across all validation sets was prioritized, since the optimization process for M/V-ATTA has been exclusively conducted on the validation set. Moreover, all the uncertainty calibration methods used in the experiments (this includes \textit{M/V-ATTA} and all the baselines) exclusively utilize the \textit{validation set} for parameter optimization, and so the \textit{training set} is only used for the DNN training process. Additionally, it is worth mentioning the ``plug and play" nature of all the methods used, since neither require any type of re-training or modification of the DNN, and therefore can be used with any previously trained model, only requiring a small validation set for parameter optimization. \\
\indent For each dataset, the experiments made with \textit{M/V-ATTA} were performed according to the following steps: using each of the eight augmentation policies (see Table \ref{aug_policies}), the learnable parameters wer optimized in the validation set, with Negative Log Likelihood (NLL) as loss function, 500 epochs, batch size of 500, Adam optimizer with a learning rate of 0.001 and all weights initialized as 1; the eight obtained versions of \textit{M/V-ATTA} were compared on the  validation set (the same used in the parameter optimization), using the Brier score, where the best performing augmentation policy was selected (see Subsection \ref{policy_validation}); using the previously found augmentation policy, the method was finally evaluated \CP{in}on the test set (see Subsection \ref{main_results}).\\
\indent It is noteworthy that test time augmentation - just like traditional augmentation - can be applied with an highly extensive set of augmentation policies thus, possibly resulting in virtually endless \CP{different} distinct scenarios. Nevertheless, for practical reasons, the experiments here presented were conducted by taking into consideration \CP{only} four \CP{different} image transformations, combining them into eight \CP{different} distinct augmentation policies. The image transformations used are: \CP{were}
\begin{itemize}
\item \textit{Flip}: a flip around the vertical axis of the image;
\item \textit{Crop}: creates a cropped input with size ratio of approximately $0.8$, extracted from a random position within the original input image;
\item \textit{Brightness}: creates, from an original input $\gamma$, a new input
$
    \gamma' = \gamma + \beta \gamma_{\text{max}}, 
$
where $\gamma_{\text{max}}$ is the maximum pixel value from $\gamma$, and $\beta$ is a random number extracted from a continuous Uniform distribution  $U(-0.5,0.5)$;
\item \textit{Contrast}: creates, from an original input $\gamma$, a new input 
$
    \gamma'' = \gamma(1 + \alpha),
$
where $\alpha$ is a random number extracted from $U (-0.2,0.2)$. \CP{Uniform distribution within the interval $[-0.2,0.2]$.}
\end{itemize}
These four types of augmentations were selected for this work based on the fact that they are common and easily replicable image transformations, thus reinforcing the merits of the proposed methods (\textit{M/V-ATTA}), by showing their applicability without the need of extensively searching a set of complex augmentations. \\
\indent The composition of the eight different augmentation policies is described in Table \ref{aug_policies}. The rationale behind the selection of such policies is based on \CP{some}the conclusions derived from the results with the preliminary approach presented in \cite{Conde_2022_BMVC}. First, all augmentation policies are composed by more than one type of augmentation, based on the fact that this was generally the best approach in \cite{Conde_2022_BMVC}. Secondly, the \textit{contrast} transformation is only present in augmentation policies that are composed by three or more types of augmentations, since there was some evidence in \cite{Conde_2022_BMVC} that this transformation had, in general, the lowest positive impact in terms of uncertainty calibration.   \\ 
\indent The subsequent subsections provide a comprehensive overview of the experiments conducted and the corresponding results of this work. To summarize, the experiments are categorized as follows:\\
 \\
\noindent \textbf{Augmentation policy validation}: Here we analyse the initial phase of the application of \textit{M/V-ATTA}. For each dataset, the methods undergo optimization and testing within the same validation set. This process aims to determine the most effective augmentation policies, for subsequent use in each specific case of the following subsections.\\
 \\
\noindent \textbf{In-distribution results}: \textit{M-ATTA} and \textit{V-ATTA}, with the parameters optimized in the previous section (each with its selected augmentation policy), are subjected to testing in the respective \textit{test sets} of each dataset. Their performance is evaluated with different uncertainty calibration-related metrics, against multiple state-of-the-art baselines. \PC{We can consider this the first of the two main subsections of experiments/results.}\\
 \\
\noindent \textbf{Robustness under distribution-shift}: The methods from the previous subsection are further compared, employing the same metrics, in two scenarios where distribution shifts in test data are present. This is made either with \textit{artificially induced} distribution-shifts or with \textit{natural} distribution-shifts. \PC{This is the second of two main subsections of experiments/results.}  \\
  \\
\noindent \textbf{Entropy on out-of-distribution samples}: \CP{Although not the main purpose of this experimental section}Additionally, the previously used methods are also compared in terms of their effects on the predictive entropy on OOD samples, under two distinct scenarios.\\
 \\
\noindent \textbf{Effects of validation set size}: Finally, we analyze how \CP{changing} the size of the validation set, in which \textit{M-ATTA} and \textit{V-ATTA} are optimized, \CP{changes}influences the performance of these methods. The relevance of these additional experiments is justified by some of the conclusions derived on Subsection \ref{main_results}, regarding over-fitting phenomena. \\
 \\

\subsection{Augmentation policy validation}
\label{policy_validation}

\begin{table*}[]
    \centering
    \begin{adjustbox}{max width=\textwidth}
    \begin{tabular}{c|c|c c c c c c c c}
    \toprule
    \multicolumn{10}{c}{{\normalsize\textbf{Augmentation Policy Validation}} } \rule{0pt}{4ex}\\
     \multicolumn{10}{c}{}\\
     \hline
     \rule{0pt}{3ex}
        ~ & ~ &  Aug. 1 & Aug. 2 & Aug. 3 & Aug. 4 & Aug. 5 & Aug. 6 & Aug. 7 & Aug. 8\\  \hline\hline
        \rule{0pt}{3ex}\multirow{2}{*}{CIFAR-10} &  M-ATTA  & 0.0399 &
0.0371 &
0.0434 &
0.0372 &
0.0376 &
0.0366 &
0.0416 &
\textbf{0.0362}  \\ 
        \rule{0pt}{2.5ex} ~ & V-ATTA  & 0.0426 &
\textbf{0.0402} &
0.0465 &
0.0417 &
0.0419 &
0.0408 &
0.0448 &
0.0408  \\ 
 \hline\hline
       \rule{0pt}{3ex}\multirow{2}{*}{CIFAR-100} & M-ATTA  & 0.1149 &
0.1126 &
0.1179 &
0.1081 &
0.1084 &
0.1109 &
0.1134 &
\textbf{0.1065}  \\ 

        \rule{0pt}{2.5ex} ~ & V-ATTA  & 0.1277 &
0.1242 &
0.1336 &
0.1246 &
0.1248 &
\textbf{0.1240} &
0.1311 &
0.1241  \\
 \hline\hline
        \rule{0pt}{3ex}\multirow{2}{*}{AID} & M-ATTA  & 0.0272 &
0.0453 &
0.0284 &
0.0254 &
0.0253 &
0.0428 &
0.0263 &
\textbf{0.0247} \\ 

        \rule{0pt}{2.5ex} ~ & V-ATTA  & 0.0316 &
0.0529 &
0.0328 &
\textbf{0.0310} &
0.0314 &
0.0507 &
0.0328 &
0.0315 \\ 
 
        \bottomrule
    \end{tabular}
    \end{adjustbox}
    \caption{Results with respect to the Brier score - in the respective validation sets of the CIFAR-10, CIFAR-100 and AID datasets - comparing the eight different augmentation policies with \CP{both our}the proposed methods (\textit{M-ATTA} and \textit{V-ATTA}). For each case, the best result is represented in bold.}
    \label{validation}
\end{table*}

In this subsection we compare the \CP{eight} augmentation policies described in Table \ref{aug_policies} in terms \CP{on how}of \textit{M-ATTA} and \textit{V-ATTA} performance by considering the Brier score. For this purpose, the experiments are done in the respective validation set of each \CP{of the three} dataset (CIFAR-10, CIFAR-100, AID). The results are presented in Table \ref{validation}. \PC{This information will be leveraged to select the best performing augmentation policy} The augmentation policies that achieve the best performance, in each case, are subsequently used \CP{in Subsection \ref{main_results}} in the experiments on \CP{done with each}the corresponding test set (Subsection \ref{main_results}).\\
\indent By analyzing the results presented in Table 2, it becomes evident that Aug. 8 consistently emerges as the most effective policy for \textit{M-ATTA}, irrespective of the dataset employed. However, this trend does not hold when evaluating \textit{V-ATTA}, where we find a different best-performing augmentation policies for each dataset. Nonetheless, the results obtained with Aug. 8 applied to \textit{V-ATTA} are still relatively close to best performing scenario, across all three datasets. We also note that Aug. 2 and Aug. 6 - which yield the best results for \textit{V-ATTA} in CIFAR-10 and CIFAR-100, repectevely - are paradoxically the least effective augmentation policies when considering the AID dataset, suggesting some dataset-dependent behaviour.\\
\indent Finally,it is noteworthy that M-ATTA consistently outperforms V-ATTA across all cases. The analogous results of those outlined in Table \ref{validation}, but evaluated on the respective test sets, can be found in Supplementary Material - ``Supplementary results".

\subsection{In-distribution results}
\label{main_results}

\begin{table*}[]
\LARGE
    \centering
    \begin{adjustbox}{max width=\textwidth}
    \begin{tabular}{c|c c c c | c c c c | c c c c }
    \toprule
    \multicolumn{13}{c}{{\textbf{In-Distribution Results}} } \rule{0pt}{4ex}\\
     \multicolumn{13}{c}{}\\
     \hline
     \rule{0pt}{3ex}
     ~ & \multicolumn{4}{c|}{CIFAR-10} & \multicolumn{4}{c|}{CIFAR-100} & \multicolumn{4}{c}{AID} \\
     \rule{0pt}{2.5ex}  ~ & Brier & ECE & mc-Brier & NLL & Brier & ECE & mc-Brier & NLL & Brier & ECE & mc-Brier & NLL \\ \hline \hline
     \rule{0pt}{3ex} 
     Vanilla & 0.0438 & 0.0367 & 0.0929 & 0.2356 & 0.1599 &  0.1325 & 0.4055 & 1.2132 & 0.0481 & 0.0306 & 0.1068 & 0.3046 \\ \hline \hline 
     \rule{0pt}{3ex} 
     T. Scaling \cite{guo2017calibration} & 0.0390 & 0.0119 & 0.0860 & 0.1909 & 0.1364  & 0.0241 & 0.3766 & 1.0208 &  {0.0493} & 0.0300 &  {0.1073} & 0.2621 \\
     \rule{0pt}{2.5ex} 
     I. Regression \cite{zadrozny2002transforming} & 0.0396 &  \textbf{ {0.0069}} & 0.0870 & 0.1918 & 0.1352 & 0.0192 & 0.3792 & 1.0471 & 0.0476 & 0.0158 &  {0.1074} &  0.2725 \\
      \rule{0pt}{2.5ex} 
      H. Binning \cite{zadrozny2001obtaining} &  {0.0462} &  0.0112 &  {0.0967} &  {0.2613} & 0.1412 & \textbf{ {0.0135}} &  0.3907 &  {1.4232} & 0.0467 & \textbf{ {0.0138}} &  {0.1083} &  {0.3376} \\
      \rule{0pt}{2.5ex} 
      Dirichlet \cite{kull2019beyond} & 0.0428  &  0.0131 & 0.0941  &  0.2265 & 0.1456 & 0.0549 &  0.4120 & 1.4220  & 0.0387  & 0.0170 & 0.0826 & 0.2070  \\
      \rule{0pt}{2.5ex} 
      Adaptive-TTA \cite{Conde_2022_BMVC} & 0.0373  &  0.0136 & 0.0819  & 0.1805  & 0.1292 & 0.0256 & 0.3645  & 1.0111  & 0.0363 & 0.0203 & 0.0823 &  0.1647 \\
     \hline \hline
     \rule{0pt}{3ex} 
     \textbf{M-ATTA} (ours) &  {0.0371} & 0.0090 &  {0.0813} &  {0.1800} &  {0.1350} & 0.0274 &  {0.3730} &  {1.0042} & \textbf{ {0.0263}} & 0.0232 & \textbf{ {0.0632}} & \textbf{ {0.1300}} \\
     \rule{0pt}{2.5ex}
     \textbf{V-ATTA} (ours) & \textbf{ {0.0358}} & 0.0130 & \textbf{ {0.0793}} & \textbf{ {0.1705}} & \textbf{ {0.1270}} & 0.0187 & \textbf{ {0.3584}} &\textbf{ {0.9565}} &  {0.0278} & 0.0221 &  {0.0645} &  {0.1365} \\ 
    
    \bottomrule
    \end{tabular}
    \end{adjustbox}
    \caption{Results with respect to four different performance metrics (Brier score, ECE, multi-class Brier score and NLL) - in the respective test sets of the CIFAR-10, CIFAR-100 and AID datasets - comparing the performance of \CP{our}the methods \textit{M-ATTA} and \textit{V-ATTA} against six baselines (Vanilla, Temperature Scaling, Isotonic Regression, Histogram Binning, Dirichlet Calibration and Adaptive-TTA). For each case, the best result is represented in bold.}
    \label{results_table}
\end{table*}

\CP{Here, we describe and discuss the results summarized in}Table \ref{results_table} shows the results obtained with both \textit{M-ATTA} and \textit{V-ATTA}, compared against the performance of \textit{temperature scaling} \cite{guo2017calibration}, \textit{isotonic regression} \cite{zadrozny2002transforming}, \textit{histogram binning} \cite{zadrozny2001obtaining}, \textit{dirichlet calibration} \cite{kull2019beyond}, \textit{Adaptive-TTA} \cite{Conde_2022_BMVC} (for details on these baseline methods see Supplementary Material - ``Baseline methods") and a \textit{vanilla} approach (referring to the results obtained from the DNN without any type of calibration method). The choice of these state-of-the-art baselines takes into consideration that these methods share, with  both \textit{M-ATTA} and \textit{V-ATTA}, one important characteristic: they can be applied with any type of classification DNN, requiring no re-training or modification of the network architecture (for this reason, all these methods can be refereed to as \textit{post-hoc} calibration methods). With the exception of \textit{dirichlet calibration}, all the baselines share a second characteristic with \textit{M/V-ATTA}: they do not alter the original classification accuracy of the DNN with which they are applied. We take the opportunity to note that \textit{dirichlet calibration} worsens the accuracy in the CIFAR-10 and CIFAR-100 datasets in 0.32\textit{pp} and 2.45\textit{pp}, respectively, and improves the accuracy in the AID in 1.3\textit{pp}. \\
\indent All methods are evaluated with the described uncertainty calibration metrics - Brier score (both ``classical" and multi-class version) and ECE (with 15 bins) - and also the widely known NLL loss. Although being a proper scoring  rule (and so, built to evaluate the quality of probabilistic predictions), the NLL loss - which is traditionally used in the training process of DNNs - cannot be considered a metric of calibration (or strong calibration) given our definition, because it only considers the quality of the predictions associated with the true class; nevertheless, given that it is a proper scoring rule and a popular metric, it is common to consider the NLL alongside to Brier score and ECE, when evaluating uncertainty calibration methods \cite{ovadia2019can, tian2021geometric, moon2020confidence}.\\
\indent We start our analysis of the results in Table \ref{results_table} by comparing all the uncertainty calibration methods against the \textit{vanilla} approach. The first main observation derived in this context is that \CP{our}the proposed methods (\textit{M/V-ATTA}), along with their predecessor \textit{Adaptive-TTA}, are the only \CP{that have}having consistently better performance than the \textit{vanilla} approach, presenting better results in all evaluation metrics and all three datasets. Contrarily, \textit{temperature scaling} worsens the performance in the AID dataset (Brier score, mc-Brier score), \textit{isotonic regression} also worsens the performance in the AID dataset (Brier score), \textit{histogram binning} worsens the performance in CIFAR-10 (Brier score, mc-Brier score, NLL), CIFAR-100 (NLL) and AID (mc-Brier score, NLL) datasets, and finally, \textit{dirichlet calibration} worsens the performance in the CIFAR-10 (mc-Brier score) and CIFAR-100 (mc-Brier score, NLL) datasets.\\
\indent Focusing now on highlighting the best results for each specific scenario, we observe that in all three datasets, one of \CP{our}the proposed methods (\textit{M-ATTA} or \textit{V-ATTA}) is the best performing method in terms of Brier score, mc-Brier score and NLL. Focusing in the three aforementioned evaluation metrics: in both CIFAR-10 and CIFAR-100 datasets, \textit{V-ATTA} is consistently the best performing method, while \textit{M-ATTA} is consistently the second or third best performing method; in the AID dataset, \textit{M-ATTA} is consistently the best performing method, while \textit{V-ATTA} is consistently the second-best performing method. On the other hand, when considering ECE, \textit{isotonic regression} is the best performing method in the CIFAR-10 dataset, while \textit{histogram binning} is best performing method in the remaining two datasets (these methods are actually biased to ECE evaluation, since they are binning-based methods, what can explain the particularly good behaviour with this metric); nonetheless, \textit{M-ATTA} and \textit{V-ATTA} still preform well when considering ECE evaluation.\\
\indent Some additional observations can be derived from Table \ref{results_table}: \textit{histogram binning} seems to be an unreliable uncertainty calibration method, given how it worsens multiple evaluation metrics in different datasets; the AID dataset is found to be an interesting case study, where three post-hoc calibration baselines struggle to accomplish consistent results; the empirical evidence presented suggests that \textit{M-ATTA} and \textit{V-ATTA} are the most consistent and generally best performing methods.\\
\indent Finally, \CP{while}although \textit{M-ATTA} consistently outperforms \textit{V-ATTA} - in terms of Brier score - when assessed in the validation sets (see Subsection \ref{policy_validation}), this superiority does not manifest comparably in the test sets. This observed disparity may imply the occurrence of over-fitting phenomena in the optimization process of \textit{M-ATTA}. It is not surprising that \textit{M-ATTA} is more prone to over-fitting phenomena than \textit{V-ATTA}, since it has a larger number of parameters (proportional to the number of classes in the respective dataset).

\subsection{Robustness under distribution-shift}

\begin{table}[]
    \centering
    \begin{adjustbox}{max width=\columnwidth}
    \begin{tabular}{c|c c c c }
    \toprule
     \multicolumn{5}{c}{{\normalsize\textbf{Distribution-Shift}} 
     (CIFAR-10 $\rightarrow$ CIFAR-10*)} \rule{0pt}{4ex}\\
     \multicolumn{5}{c}{}\\
     \hline
     \rule{0pt}{2.5ex}  ~ & Brier & ECE & mc-Brier & NLL  \\ \hline \hline
     \rule{0pt}{3ex} 
     Vanilla & 0.3083 & 0.3132 & 0.7012 & 2.3530  \\ \hline \hline 
     \rule{0pt}{3ex} 
     T. Scaling \cite{guo2017calibration} & 0.2452 & 0.2239 & 0.6258  & 1.5969  \\
     \rule{0pt}{2.5ex} 
     I. Regression \cite{zadrozny2002transforming} & 0.2282 & \textbf{0.1998}  & 0.6175 & 1.5796 \\
      \rule{0pt}{2.5ex} 
      H. Binning \cite{zadrozny2001obtaining} & 0.2931  & 0.2807 &  0.6994 &  1.9913  \\
      \rule{0pt}{2.5ex} 
      Dirichlet \cite{kull2019beyond} & 0.2668  & 0.2529  & 0.6704  & 2.4788  \\
      \rule{0pt}{2.5ex} 
      Adaptive-TTA \cite{Conde_2022_BMVC} & 0.2520  & 0.2366  & 0.6362  & 1.9094  \\
     \hline \hline
     \rule{0pt}{3ex} 
     \textbf{M-ATTA} (ours) & \textbf{0.2224}  & 0.2058 & \textbf{0.5963}  & 1.5537    \\
     \rule{0pt}{2.5ex}
     \textbf{V-ATTA} (ours) & 0.2264 & 0.2054 & 0.6015 & \textbf{1.5301}  \\ 
    
    \bottomrule
    \end{tabular}
    \end{adjustbox}
    \caption{Results with respect to the distribution-shift scenario created by injecting \textit{Gaussian noise} and \textit{Gaussian blur} in the test set of the CIFAR-10 dataset. The metrics and baselines used are the same as before.}
    \label{datashift_1}
\end{table}

\begin{table}[]
    \centering
    \begin{adjustbox}{max width=\columnwidth}
    \begin{tabular}{c|c c c c }
    \toprule
     \multicolumn{5}{c}{{\normalsize\textbf{Distribution-Shift}} 
     (AID $\rightarrow$ UC-Merced)} \rule{0pt}{4ex}\\
     \multicolumn{5}{c}{}\\
     \hline
     \rule{0pt}{2.5ex}  ~ & Brier & ECE & mc-Brier & NLL  \\ \hline \hline
     \rule{0pt}{3ex} 
     Vanilla & 0.3170 & 0.3064 & 0.7469 & 3.2490  \\ \hline \hline 
     \rule{0pt}{3ex} 
     T. Scaling \cite{guo2017calibration} & 0.2624 & 0.2000 & 0.6848  & 2.2624  \\
     \rule{0pt}{2.5ex} 
     I. Regression \cite{zadrozny2002transforming} & 0.2752 & 0.2393  & 0.7085 & 2.3293 \\
      \rule{0pt}{2.5ex} 
      H. Binning \cite{zadrozny2001obtaining} & 0.2893  & 0.2596 &  0.7310 &  2.6855  \\
      \rule{0pt}{2.5ex} 
      Dirichlet \cite{kull2019beyond} & 0.2793 & 0.2580 &  \textbf{0.6666} &  2.5875  \\
      \rule{0pt}{2.5ex} 
      Adaptive-TTA \cite{Conde_2022_BMVC} & 0.2788  & 0.2549  & 0.7009  & 2.5609  \\
     \hline \hline
     \rule{0pt}{3ex} 
     \textbf{M-ATTA} (ours) & 0.2657  & 0.2058 & 0.6870  & 2.3033    \\
     \rule{0pt}{2.5ex}
     \textbf{V-ATTA} (ours) & \textbf{0.2519} & \textbf{0.1897} & 0.6726 & \textbf{2.1471}  \\ 
    
    \bottomrule
    \end{tabular}
    \end{adjustbox}
    \caption{Results with respect to the distribution-shift scenario created by substituting the AID test set, for the subset of the UC-Merced dataset with homologous classes. The metrics and baselines \CP{used} are the same as before. \CP{those of Table \ref{results_table}.}}
    \label{datashift_2}
\end{table}

\indent In \CP{the present}this subsection, it is evaluated the ability of the proposed methods (once again, against the \CP{different} baselines) to improve uncertainty calibration but, this time, \CP{ even when} when the distribution of data \CP{in}on the test set differs significantly from that of the training and validation sets. \CP{These type of shifts in the distribution of} Distribution-shifts on the test data has been shown to have negative effects on uncertainty calibration, even when applying state-of-the-art calibration methods \cite{ovadia2019can}. Furthermore, evaluating uncertainty calibration \CP{in such setups}subjected to distribution-shift is - in our perspective - critical when considering the reliability of DNNs, since distribution shifts are ubiquitous in practical real-world contexts. For the purpose of these experiments, \CP{there are created two different}two distinct \textit{distribution-shift scenarios} have been considered.\\
\indent The first scenario - whose results are presented in Table \ref{datashift_1} - is created by inducing, in the CIFAR-10 test set, \textit{artificial} shifts in distribution, through the injection of \textit{Gaussian noise} (mean of 0, variance taken from $U(10,50)$ for each input, independently for each channel) and \textit{Gaussian blur} (blurs the input image using a Gaussian filter with a random kernel size between 1 and 5). These type of corruptions and perturbations to the test set have been shown to have a harmful effect on DNN performance in classification problems \cite{hendrycks2019benchmarking}.  \\
\indent The second scenario  - whose results are presented in Table \ref{datashift_2} - is created by replacing the AID test set, with a subset of the \textit{UC Merced Land Use Dataset} \cite{yang2010bag} (UC-Merced), consisting of the images whose classes are homologous to those of the AID dataset. Specifically, there are 10 classes in common between the AID and the UC-Merced datasets; we simply select the images of the UC-Merced dataset with those classes, creating the new shifted test set, with 1000 images in total.\\
\indent Regarding the results presented in Tables \ref{datashift_1} and \ref{datashift_2}, we start by observing that the distribution-shifts, induced by the new test sets, severely worsen the uncertainty calibration of the models, verified by a substantial increase of the values associated with \CP{any of} the evaluation metrics. Nonetheless, in these scenarios - and contrarily to what happened in the \textit{in-distribution} scenarios - all the methods evaluated consistently show positive effects across the various metrics (with exception to \textit{dirichlet calibration} \CP{, when evaluated with NLL in the first distribution-shift scenario}); this may hint that the \textit{post-hoc} calibration baselines have a more consistent behaviour when the severity of miscalibration is higher. Moreover, it is again observable that \textit{M-ATTA} and \textit{V-ATTA} are the overall best performing methods in these new experimental setups.

\subsection{Entropy on OOD samples}

\begin{figure}[]
    \centering
    \includegraphics[width=1\columnwidth]{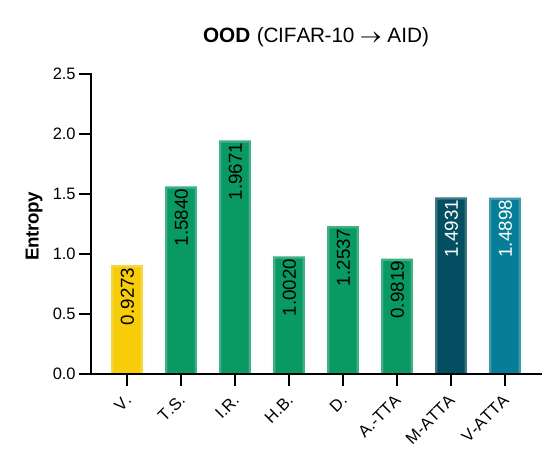}  
    \caption{Results with respect to the average OOD prediction entropy, using the DNN and the calibration methods that have been trained/optimized on the CIFAR-10, exposed to the AID test set.}
    \label{fig_OOD_1}
\end{figure}

\begin{figure}[]
    \centering
    \includegraphics[width=1\columnwidth]{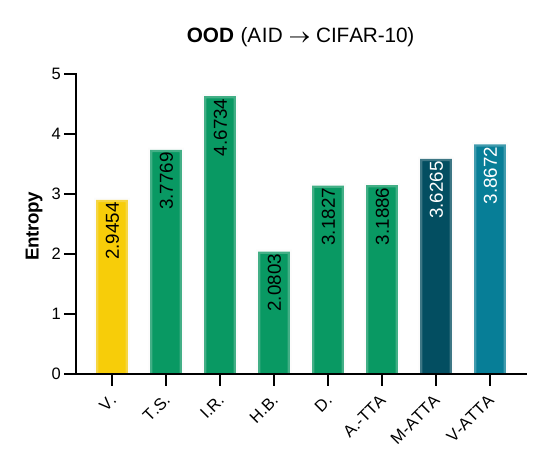}  
    \caption{Results with respect to the average OOD prediction entropy, using the DNN and the calibration methods\CP{(those also evaluated in the previous subsections)} that have been \CP{are} trained/optimized \CP{in}on the AID, exposed to the CIFAR-10 test set.}
    \label{fig_OOD_2}
\end{figure}

\CP{Although not the main purpose of the experimental section of this work, we are also interested in} Additionally, we think it is beneficial to understand how the previously evaluated methods affect the prediction vector of the DNN, when exposed to OOD samples. The expected behaviour \CP{that is expected}, from a reliable model, is that the elements of the prediction vector are as close as possible to being uniformly distributed \textit{i.e.}, the prediction vector has the highest possible entropy. \\
\indent For the purpose of these experiments, we create two OOD scenarios: first, by exposing the DNN and the calibration methods, that were trained/optimized in CIFAR-10, to the AID test set; secondly, by exposing the DNN and the calibration methods, that were trained/optimized in the AID, to the CIFAR-10 test set. Recall that the \textit{entropy}, for a vector $\mathbf{p}=(p_1,p_2,\ldots,p_k) \in \mathbb{R}^k$, is computed as 
\begin{align}
    \text{Entropy}(\mathbf{p}) = - \sum_{i=1}^k \frac{p_i \log_2(p_i)}{k}.
\end{align}
In Figures \ref{fig_OOD_1} and \ref{fig_OOD_2}, the entropy values are presented as the average entropy of all the prediction vectors \CP{, of} obtained on the respective test set. As previously referred - and contrarily to what is common when measuring uncertainty calibration - we expect the entropy value to be the highest possible.\\
\indent Regarding the \CP{concrete} results presented in the Figures \ref{fig_OOD_1} and \ref{fig_OOD_2}, two main observations can be derived: \textit{M-ATTA},\textit{V-ATTA} and \textit{temperature scaling} have a similar effect on ODD prediction entropy, showing overall good results; nonetheless, \textit{isotonic regression} is the best performing method in both of the evaluated scenarios, in the context of OOD prediction entropy.

\subsection{Effects of validation set size}

\begin{figure*}[]
    \centering
    \includegraphics[width=1\textwidth]{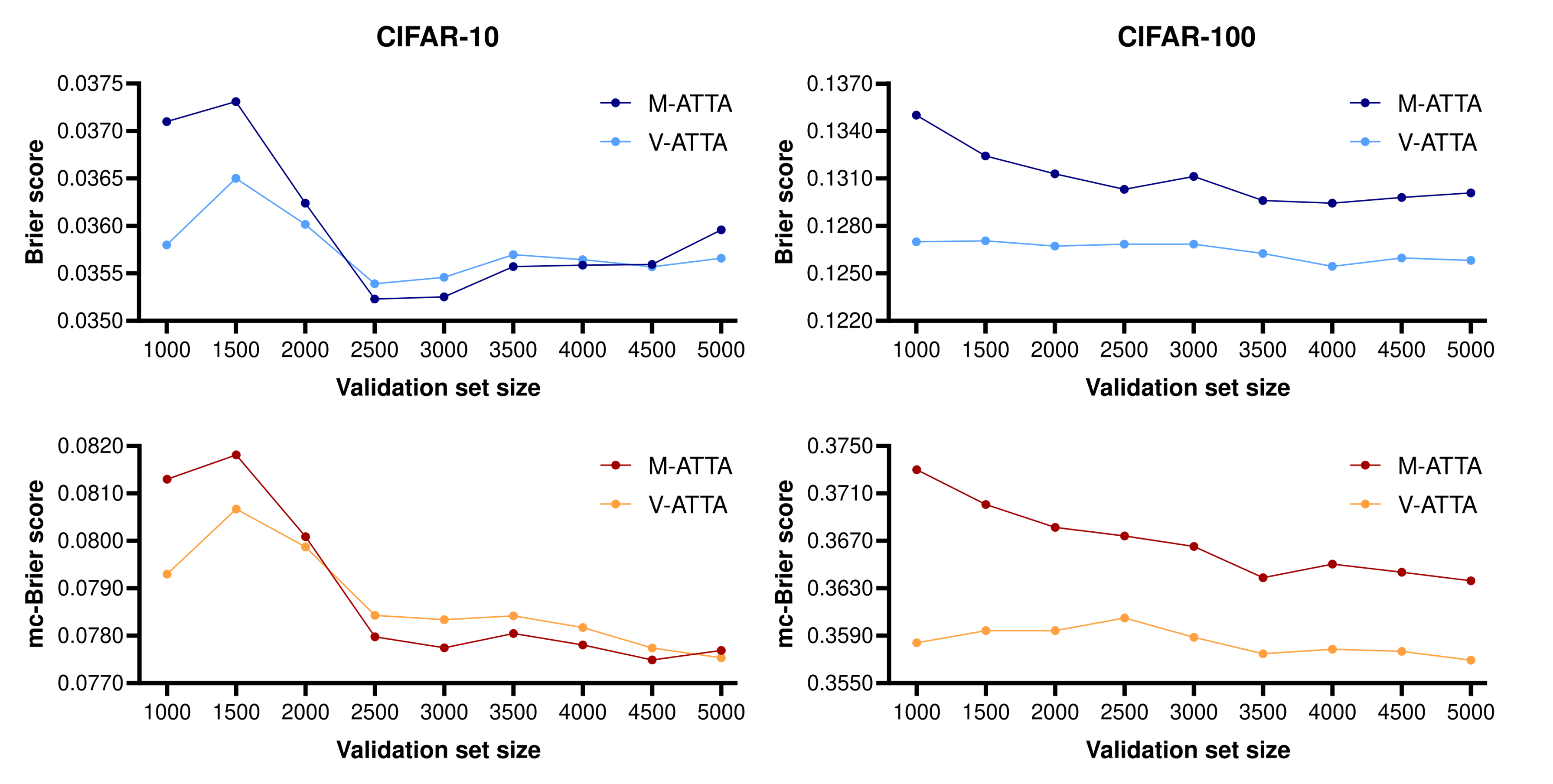}  
    \caption{Comparing the Brier and mc-Brier scores obtained with \textit{M-ATTA} and \textit{V-ATTA}, with different validation/test set ratios, evaluated in the respective test sets of both CIFAR-10 and CIFAR-100 datasets.}
    \label{fig_val}
\end{figure*}

In the previous subsection we observed that, when considering Brier score, mc-Brier score and NLL, \textit{V-ATTA} is the best performing method in CIFAR-10 and CIFAR-100 (that are datasets with similar characteristics) while \textit{M-ATTA} is the best performing method \CP{in}on the AID dataset. While this is possibly justified by the \CP{different} nature of the AID dataset (being substantially different from CIFAR-10/100), we explore the possibility that this difference in behavior is caused by \CP{different proportions in terms}the size of the validation/test sets. Furthermore, the differences witnessed between Tables \ref{validation} and \ref{results_table}, in terms of Brier score, suggest the possibility of a partial over-fitting phenomena. As such, this subsection is centered around experimenting \CP{with the} variations on the size ratio between the validation and the test set on the CIFAR-10 and CIFAR-100 datasets, and understanding their impact in the performance of both \textit{M-ATTA} and \textit{V-ATTA}\\
\indent The results are shown in Figure \ref{fig_val} (analogous results for ECE and NLL are presented in Supplementary Material - ``Supplementary results"). In these experiments, both \textit{M-ATTA} and \textit{V-ATTA} are optimized and tested with different validation/test set size ratios, beginning with 1000 validation set samples (like in Subsection \ref{main_results}) and iteratively adding 500 samples (while subtracting them from the test set) until reaching 5000 (this is done in both CIFAR-10 and CIFAR-100 datasets). In the CIFAR-10 dataset, we observe that \textit{M-ATTA} actually outperforms \textit{V-ATTA} when the validation set size is around the range between 2500 and 4000 (in this range the validation set size is approximately 1/3 of the test set size, just like in the AID setup for the previous experiments). On the other hand, in the CIFAR-100 experiments, although some improvement in performance is observed with \textit{M-ATTA}, it is not enough to outperform \textit{V-ATTA}. It is somewhat expected that the problem of over-fitting is more easily addressed when working with CIFAR-10, since the \textit{M-ATTA} model used in CIFAR-100 has ten times more parameters than the one used with CIFAR-10.

\section{Final Remarks}

Based on the results presented and discussed in the previous section\PC{, complemented by the supplementary material}, we highlight \CP{some the} the key conclusions that can be derived from this work:
\begin{itemize}
    \item \textit{M-ATTA}, \textit{V-ATTA} and \textit{Adaptive-TTA} are the most robust uncertainty calibration methods, since - contrarily to the other baseline methods \CP{selected for comparison} - they never produce worse results (compared to the \textit{vanilla} approach) with any of the selected evaluation metrics, independently of the experimental scenario. 
    \item \textit{M-ATTA} and \textit{V-ATTA} can be considered the overall best-performing methods, either in an \textit{in-distribution} setup or under \textit{distribution-shift}, since they generally outperform most - or all - baselines in terms of Brier score, mc-Brier score and NLL. This is complemented with a \CP{nonetheless robust}solid performance in terms of ECE.
    \item \CP{Our}The proposed methods generally lose to either \textit{isotonic regression} or \textit{histogram binning} when evaluated with the ECE. These baseline methods are somewhat biased towards ECE evaluation, since they leverage binning for obtaining better calibrated predictions. We speculate this justifies the inconsistency of these methods with the other evaluation metrics and the specially good performance with the ECE.
    \item Besides their performance and robustness in terms of uncertainty calibration, \textit{M-ATTA} and \textit{V-ATTA} show also positive effects on OOD predictive entropy. Nonetheless, \textit{isotonic regression} outperforms our methods in this \CP{type of} evaluation.
    \item It is possible to observe a partial over-fitting phenomena with \textit{M-ATTA}, when evaluated in CIFAR-10/100 (\textit{in-distribution}). We demonstrate that the size of the validation set significantly influences this phenomenon, particularly in scenarios with a low number of classes (as in CIFAR-10); however, its significance appears to diminish with a higher number of classes (as in CIFAR-100).
\end{itemize}
\indent We care to note that, besides the advantages presented by the proposed methods (\textit{M/V-ATTA}), this comes at the cost of higher inference time. This inference time will be proportional to the number of augmentations used, since the DNN will make a prediction for each augmented input.\\
\indent For future work, we identify some challenges that can be addressed: studying different solutions for the partial over-fitting phenomena found with \textit{M-ATTA}; exploring more complex types of augmentations; adapting our methods and extending their empirical evaluation to scenarios of object detection and/or semantic segmentation.

\appendices


\ifCLASSOPTIONcompsoc
  \section*{Acknowledgments}
\else
  \section*{Acknowledgment}
\fi

This work has been supported by the Portuguese Foundation for Science  and  Technology (FCT), via the project $GreenBotics$ (PTDC/EEI-ROB/2459/2021), and partially by Critical Software, SA.

\ifCLASSOPTIONcaptionsoff
  \newpage
\fi



%


\bibliographystyle{abbrv}
\bibliography{ref}


%




\end{document}


\maketitle

\section*{\textit{M/V-ATTA} accuracy consistency}

In this supplementary section we present the \CP{formal} proof that \textit{M/V-ATTA} does not alter the accuracy of the model with which the proposed method is applied. The proof is analogous for \textit{M-ATTA} and \textit{V-ATTA}, but both are presented \CP{nonetheless}for sake of clarity. We start by \CP{stating} a \CP{formal} definition for the \CP{trivial concept of} \textit{accuracy} of a classification model on a given set.\CP{, that will help in terms of notation for the proof of the referred result.}

\begin{defi}
    For a set of ordered inputs $X=\{x_1,\ldots,x_n\}$, \CP{respective}with ground-truth classes denoted by $Y=\{y_1,\ldots,y_n\}$ (for a problem with $k$ \CP{different} classes), \CP{ and a model $f: X \rightarrow \Delta_k$, with associated set of outputs } the set of outputs of a model $f: X \rightarrow \Delta_k$ is given by
    \begin{align}
        Y' = \left\{  \argmax_{i \in \{1,\ldots,k\}} f(x_1), \ldots,  \argmax_{i \in \{1,\ldots,k\}} f(x_n) \right\} = \{y'_1,\ldots,y'_n\}.
    \end{align}
    Hence, we can define the accuracy of the model $f$ with respect to $(X,Y)$ as
    \begin{align}
        acc\left(f(X),Y\right) = \sum_{i=1}^n \frac{\delta(y'_i,y_i)}{n},
    \end{align}
    where
    \begin{align}
       \delta(a,b) = 
       \begin{cases}
           1, \quad \text{if} \ a=b \\
           0, \quad \text{if} \ a \neq b
       \end{cases}.
    \end{align}
\end{defi}
Based on the above, \CP{Now} we can \CP{finally} consider the following propositions.

\begin{prop}
    \textit{M-ATTA} does not alter the accuracy of the model with which is applied.
\end{prop}
\begin{proof}
Let us consider a set of ordered inputs $X=\{x_1,\ldots,x_n\}$, respective ground truth classes $Y=\{y_1,\ldots,y_n\}$ (for a problem with $k$ different classes) and a model $f: X \rightarrow \Delta_k$, with associated set of outputs 
    \begin{align}
        Y' = \left\{  \argmax_{i \in \{1,\ldots,k\}} f(x_1), \ldots,  \argmax_{i \in \{1,\ldots,k\}} f(x_n) \right\} = \{y'_1,\ldots,y'_n\}.
    \end{align}    
We can now take
 $h: X \rightarrow \Delta_k$, that represents the model obtained by applying $f$ with \textit{M-ATTA}, and the respective set of outputs
    \begin{align}
        Y'' = \left\{  \argmax_{i \in \{1,\ldots,k\}} h(x_1), \ldots,  \argmax_{i \in \{1,\ldots,k\}} h(x_n) \right\} = \{y''_1,\ldots,y''_n\}.
    \end{align}
    \\
\indent Let us consider a \CP{generic}given element $x_i \in X$, and take $\mathbf{p}^0 = f(x_i)$. As such,
    \begin{align}
        \mathbf{p}\left(\tilde{\omega}\right) = (1-\tilde{\omega})\mathbf{p}^0 + \tilde{\omega} \text{ } \sigma(\mathbf{W} \odot \mathbf{Z}) I_m = h(x_i),
    \end{align}
    with
    \begin{align}
    \label{omega}
    \tilde{\omega} = \max \Big{\{} \omega \in [0,\omega^*] : \argmax_{i \in \{1,\ldots,k\}} \mathbf{p}\left(\tilde{\omega}\right) = \argmax_{i \in \{1,\ldots,k\}} \mathbf{p}^0 \Big{\}}.
    \end{align}
We observe that $\sigma$, $\mathbf{W}$, $\mathbf{Z}$ and $I_m$ have been defined in the main document.\\
\indent From equation \ref{omega}, we can infer 
\begin{align}
    \argmax_{i \in \{1,\ldots,k\}} \mathbf{p}\left(\tilde{\omega}\right) = \argmax_{i \in \{1,\ldots,k\}} \mathbf{p}^0 \Leftrightarrow 
    \argmax_{i \in \{1,\ldots,k\}} h(x_i) = \argmax_{i \in \{1,\ldots,k\}} f(x_i)
    \Leftrightarrow
    y''_i = y'_i.
\end{align}
Since this is true for a generic $x_i \in X$, then $Y''=Y'$. As such, we finally \CP{conclude}show that
\begin{align}
    acc\left(f(X),Y\right) = 
    \sum_{i=1}^n \frac{\delta(y'_i,y_i)}{n} =  \sum_{i=1}^n \frac{\delta(y''_i,y_i)}{n} = 
    acc\left(h(X),Y\right),
\end{align}
which concludes the proof.
\end{proof}

\begin{prop}
    \textit{V-ATTA} does not alter the accuracy of the model with which is applied.
\end{prop}
\begin{proof}
\CP{Let us}Given a set of ordered inputs $X=\{x_1,\ldots,x_n\}$, \CP{respective}the ground-truth \CP{classes} $Y=\{y_1,\ldots,y_n\}$ (for a problem with $k$ \CP{different} classes), the \CP{and a} model $f: X \rightarrow \Delta_k$ \CP{, with associated}yields a set of outputs 
    \begin{align}
        Y' = \left\{  \argmax_{i \in \{1,\ldots,k\}} f(x_1), \ldots,  \argmax_{i \in \{1,\ldots,k\}} f(x_n) \right\} = \{y'_1,\ldots,y'_n\}.
    \end{align}    
Now, consider \CP{We can now take} $l: X \rightarrow \Delta_k$ \CP{that represents} the model obtained by applying $f$ with \textit{V-ATTA}, with the outputs given by
    \begin{align}
        Y'' = \left\{  \argmax_{i \in \{1,\ldots,k\}} l(x_1), \ldots,  \argmax_{i \in \{1,\ldots,k\}} l(x_n) \right\} = \{y''_1,\ldots,y''_n\}.
    \end{align}
    \\
\indent Let us consider a \CP{generic}given element $x_i \in X$, and take $\mathbf{p}^0 = f(x_i)$. As such,
    \begin{align}
    \mathbf{p}\left(\tilde{\omega}\right) = 
    (1-\tilde{\omega})\mathbf{p}^0 + \tilde{\omega} \text{ } \sigma (\mathbf{W}_d  \mathbf{Z}^{\text{T}}) I_m = l(x_i),
\end{align}
    with
    \begin{align}
    \label{omega_2}
    \tilde{\omega} = \max \Big{\{} \omega \in [0,\omega^*] : \argmax_{i \in \{1,\ldots,k\}} \mathbf{p}\left(\tilde{\omega}\right) = \argmax_{i \in \{1,\ldots,k\}} \mathbf{p}^0 \Big{\}}.
    \end{align}
Let us note that $\mathbf{W}_d$ is defined in the main document.\\
\indent From equation \ref{omega_2}, we can infer 
\begin{align}
    \argmax_{i \in \{1,\ldots,k\}} \mathbf{p}\left(\tilde{\omega}\right) = \argmax_{i \in \{1,\ldots,k\}} \mathbf{p}^0 \Leftrightarrow 
    \argmax_{i \in \{1,\ldots,k\}} l(x_i) = \argmax_{i \in \{1,\ldots,k\}} f(x_i)
    \Leftrightarrow
    y''_i = y'_i.
\end{align}
Since this is true for any $x_i \in X$, then $Y''=Y'$. As such, we finally conclude that
\begin{align}
    acc\left(f(X),Y\right) = 
    \sum_{i=1}^n \frac{\delta(y'_i,y_i)}{n} =  \sum_{i=1}^n \frac{\delta(y''_i,y_i)}{n} = 
    acc\left(l(X),Y\right),
\end{align}
which therefore concludes the proof.
\end{proof}

\section*{Supplementary results}

\begin{figure*}[h]
    \centering
    \includegraphics[width=1\textwidth]{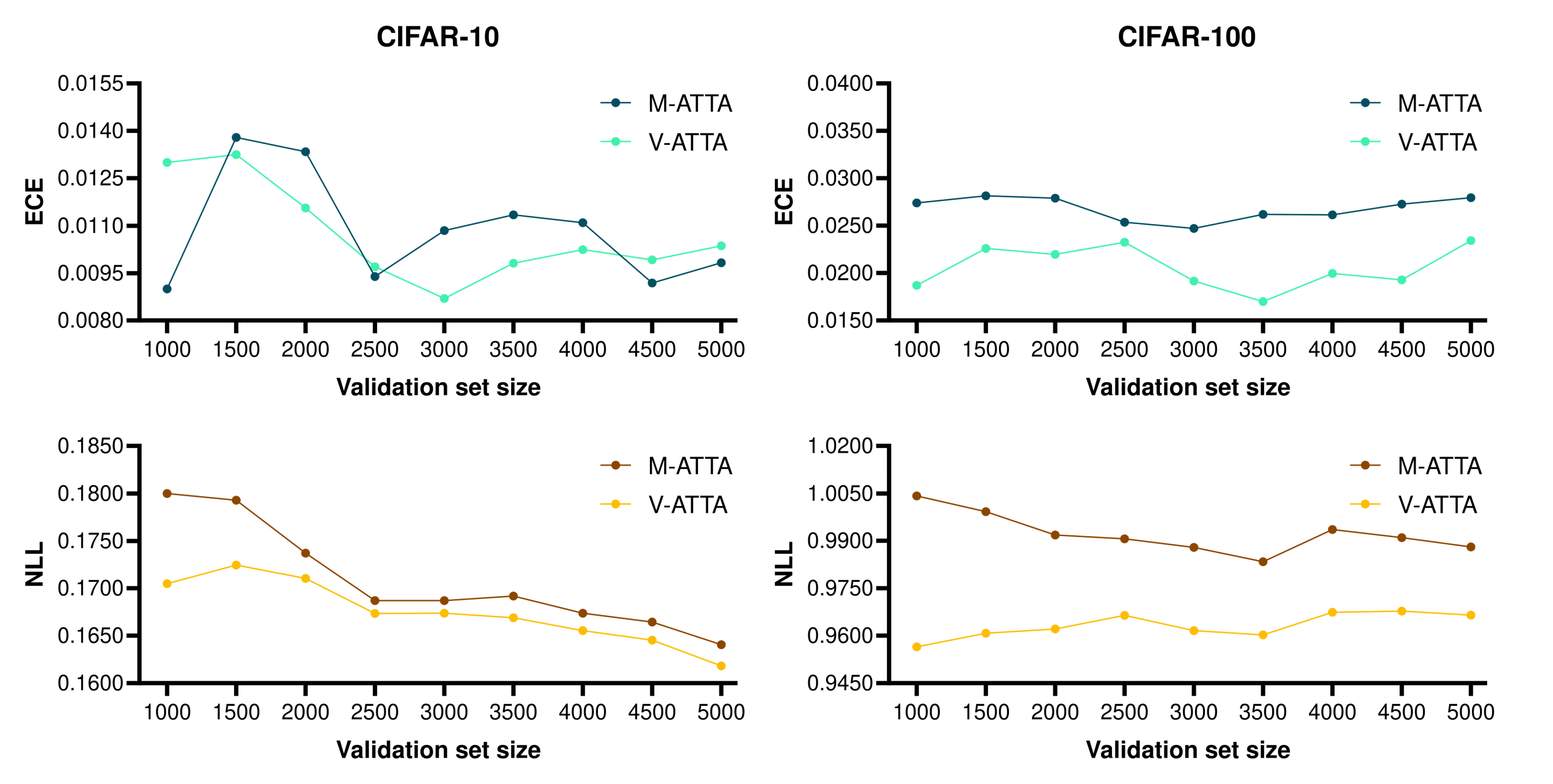}  
    \caption{Comparing the ECE and NLL scores obtained with \textit{M-ATTA} and \textit{V-ATTA}, with different validation/test set ratios, evaluated in the respective test sets of both CIFAR-10 and CIFAR-100 datasets.}
    \label{fig_val_2}
\end{figure*}

Table \ref{table_aug_2} and Figure \ref{fig_val_2} present supplementary results to complement those presented and discussed in the main document.

\begin{table}[h]
    \centering
    \begin{adjustbox}{width=\textwidth}
    \begin{tabular}{c|c|c c c c c c c c}
    \toprule
        ~ & ~ & Aug. 1 & Aug. 2 & Aug. 3 & Aug. 4 & Aug. 5 & Aug. 6 & Aug. 7 & Aug. 8\\  \hline\hline
        \rule{0pt}{3ex}\multirow{2}{*}{CIFAR-10} &  M-ATTA   & \textbf{0.0360} &
0.0362 &
0.0387 &
0.0365 &
0.0363 &
0.0369 &
0.0393 &
0.0371  \\ 
 
        \rule{0pt}{3ex} ~ & V-ATTA & 0.0358 &
0.0358 &
0.0386 &
0.0354 &
\textbf{0.0352} &
0.0359 &
0.0381 &
0.0357  \\

         \hline\hline
       \rule{0pt}{3ex}\multirow{2}{*}{CIFAR-100} & M-ATTA  & 0.1365 &
\textbf{0.1335} &
0.1422 &
0.1343 &
0.1343 &
0.1339 &
0.1410 &
0.1350  \\ 
    
        \rule{0pt}{3ex} ~ & V-ATTA  & 0.1308 &
0.1271 &
0.1359 &
0.1272 &
0.1271 &
0.1270 &
0.1334 &
\textbf{0.1267}  \\

         \hline\hline
        \rule{0pt}{3ex}\multirow{2}{*}{AID} & M-ATTA   & 0.0281 &
0.0461 &
0.0277 &
0.0263 &
\textbf{0.0259} &
0.0433 &
0.0260 &
0.0263 \\ 
    
        \rule{0pt}{3ex} ~ & V-ATTA   & 0.0296 &
0.0520 &
0.0290 &
0.0278 &
0.0276 &
0.0490 &
0.0272 &
\textbf{0.0271} \\

        \bottomrule
    \end{tabular}
    \end{adjustbox}
        \caption{Results \CP{with respect to}in terms of the Brier score - in the respective test sets of the CIFAR-10, CIFAR-100 and AID datasets - comparing the \CP{eight different} augmentation policies with both \CP{our} methods (\textit{M-ATTA} and \textit{V-ATTA}). For each case, the best result is represented in bold.}
        \label{table_aug_2}
\end{table}

\section*{Baseline methods}

The baseline methods used in the experiments are \textit{temperature scaling} \cite{guo2017calibration}, \textit{isotonic regression} \cite{zadrozny2002transforming}, \textit{histogram binning} \cite{zadrozny2001obtaining}, \textit{dirichlet calibration} \cite{kull2019beyond} and \textit{Adaptive-TTA} \cite{Conde_2022_BMVC}. We describe, in this supplementary section, the details regarding \textit{temperature scaling}, \textit{isotonic regression} and \textit{histogram binning}. As for \textit{dirichlet calibration} and \textit{Adaptive-TTA}, given their higher complexity, we refer to the original papers for further details. Nonetheless, it is important to make two observations regarding the application of both these baselines: \textit{dirichlet calibration} is applied with ODIR regularization, since this is recommended by the authors when working with DNNs; \textit{Adaptive-TTA} is applied with an augmentation policy that leverages \textit{flip}, \textit{crop} and \textit{brightness} transformations, based on the conclusions and recommendations taken from the original paper.


\subsection*{Temperature scaling \cite{guo2017calibration}}

\indent Let us consider, for a supervised problem with $k$ different classes, the aforementioned pair of random variables $(X,Y)$ and a model $f: X \rightarrow \Delta_k$. Let us now take $g: X \rightarrow \mathbb{R}^k$ such that
\begin{align}
\label{g}
    f := g \circ \sigma , 
\end{align}
where $\sigma: \mathbb{R}^k \rightarrow \Delta_k$ is the \textit{softmax} function. The output vectors of $g$ are usually called \textit{logit vectors} and each element of such vector is considered a \textit{logit}. \\
When using \textit{temperature scaling}, we replace the usual \textit{softmax} function with the \textit{temperature scaled} version $\sigma_{TS}: \mathbb{R}^k \rightarrow \Delta_k$. This new function is applied, for $T>0$, in the following way
\begin{align}
    \sigma_{TS}(\mathbf{z})_{\{i\}} = \frac{e^{z_i/T}}{\sum_{j=1}^k e^{z_j/T}} , \qquad i=1,2,\ldots,k.
\end{align}
The parameter $T$ is optimized \wrt  the NLL in a given validation set. This way, we produce a new \textit{temperature scaled} model $f_{TS}: X \rightarrow \Delta_k$ defined as
\begin{align}
    f_{TS} := g \circ \sigma_{TS}.
\end{align}
\indent Finally, we note that \textit{temperature scaling} does not change the accuracy of the model with which it is applied.
\PC{\begin{prop}
    Temperature scaling does not change the accuracy of the model in which it is applied.
\end{prop} 
\textbf{Proof}: We want to proof that the accuracy of $f$ and $f_{TS}$ is the same. Let us consider $(x,y)$ a realization of $(X,Y)$ and a logit vector $\mathbf{z} \in \mathbb{R}^k$, such that 
\begin{align}
    g(x) = \mathbf{z}.
\end{align}
Let us take $i_1,i_2 \in \{1,\ldots,k\}$, such that $i_1 \neq i_2$ and 
\begin{align}
\label{ts:1}
    \sigma_{SM}(\mathbf{z})_{\{i_1\}} \leq \sigma_{SM}(\mathbf{z})_{\{i_2\}}.
\end{align}
From \eqref{ts:1} and since $T>0$, the following implications hold true
\begin{align*}
    \frac{e^{z_{i_1}}}{\sum_{j=1}^k e^{z_j}} \leq
    \frac{e^{z_{i_2}}}{\sum_{j=1}^k e^{z_j}} & \implies
    e^{z_{i_1}} \leq e^{z_{i_2}} \\ & \implies 
    e^{z_{i_1}/T} \leq e^{z_{i_2}/T} \\ & \implies
    \frac{e^{z_{i_1}/T}}{\sum_{j=1}^k e^{z_j/T}} \leq
    \frac{e^{z_{i_2}/T}}{\sum_{j=1}^k e^{z_j/T}} 
\PC{
    \frac{1}{\sigma_{SM}(\mathbf{z})_{i_1}} \geq \frac{1}{\sigma_{SM}(\mathbf{z})_{i_2}} \implies 
    \left( \frac{\sum_{j=1}^k e^{z_j}}{e^{z_{i_1}}}\right)^{1/T} \geq
    \left( \frac{\sum_{j=1}^k e^{z_j}}{e^{z_{i_2}}}\right)^{1/T} \implies \\
    \bigskip
    \\
    \implies
    \frac{\sum_{j=1}^k e^{z_j/T}}{e^{z_{i_1}/T}} \geq
    \frac{\sum_{j=1}^k e^{z_j/T}}{e^{z_{i_2}/T}} \implies
    \frac{e^{z_{i_1}/T}}{\sum_{j=1}^k e^{z_j/T}} \leq
    \frac{e^{z_{i_2}/T}}{\sum_{j=1}^k e^{z_j/T}}
    }
\end{align*}
and thus
\begin{align}
\label{ts:2}
    \sigma_{TS}(\mathbf{z})_{\{i_1\}} \leq \sigma_{TS}(\mathbf{z})_{\{i_2\}},
\end{align}
which implicates that
\begin{align}
    \max_{i \in \{1,\ldots,k\}} \sigma_{SM}(\mathbf{z}) = \max_{i \in \{1,\ldots,k\}} \sigma_{TS}(\mathbf{z}).
\end{align}
Therefore, the class predicted by $f_{TS}$ will be the same as predicted by $f$, which means that \textit{temperature scaling} will not change the model's accuracy. $\square$}

\subsection*{Isotonic regression \cite{zadrozny2002transforming}}

 For a problem with $k$ classes and $N$ different realizations of $(X,Y)$,
let us consider all the $p^j$ predictions ($j \in [1,k N]$) given by the model $f: X \rightarrow \Delta_k$, for each realization.
Let us now take, for some $M \in \mathbb{N}$, the sets of real values
\begin{align}
   a = \{ a_0, a_1, \ldots, a_M \}, \quad \theta = \{ \theta_1, \theta_2, \ldots, \theta_M \},
\end{align}
where $\theta_i \in [0,1], \forall i \in \{1,\ldots,M\}$.\\
\indent \textit{Isotonic regression} is a non-parametric method that replaces each prediction $p^j$ with a new prediction
\begin{align}
\begin{cases}
        & \theta_i, \qquad \text{if } \quad a_{i-1}< p^j \leq a_i, \\
    & \theta_1, \qquad \text{if } \quad p^j = 0.
\end{cases}
\end{align}
$M$, $a$ and $\theta$ are obtained from the following optimization problem
\begin{align}
    & \min_{M,a,\theta} \sum_{i=1}^M \sum_{j=1}^N \mathbbold{1}_{(a_{i-1}, a_i]}(p^j) (\theta_i - o^j)^2 +
    \mathbbold{1}_{\{0 \}}(p^j) (\theta_1 - o^j)^2, \\
    & \text{subject to } \quad  0=a_0<a_1<\ldots<a_M=1, \\
     & \qquad \qquad \qquad \theta_1 \leq \theta_2 \leq \ldots \leq \theta_M.
\end{align}
where $\mathbbold{1}$ is the \textit{indicator function} and $o^j$ equals 1 if the true class corresponds to the prediction and 0 otherwise.\\
\indent In the context of classification, cases where the maximum confidence value of the prediction probability vector shares the same bin with another prediction from such vector, can be trivially solved by considering the initial maximum for the prediction class. As such, just like \textit{temperature scaling}, \textit{isotonic regression} does not alter the accuracy of the model in which it is applied.

\subsection*{Histogram binning \cite{zadrozny2001obtaining}}

\textit{Histogram binning} is also a non-parametric method and can be seen as a stricter version of \textit{isotonic regression}. In this case, we create a set of bins $\{ B_1, B_2, \ldots, B_M \}$ with boundaries
\begin{align}
    0=a_0<a_1<\ldots<a_M=1
\end{align}
such that, for each prediction $p^j$
\begin{align}
\begin{cases}
        & p^j \in B_i, \qquad \text{if } \quad a_{i-1}< p^j \leq a_i, \\
    & p^j \in B_1, \qquad \text{if } \quad p^j = 0.
\end{cases}
\end{align}
After defining the binning scheme, the \textit{histogram binning} method replaces each prediction $p^j$ with a new prediction
\begin{align}
\begin{cases}
        & acc(B_i), \qquad \text{if } \quad a_{i-1}< p^j \leq a_i, \\
    & acc(B_1), \qquad \text{if } \quad p^j = 0,
\end{cases}
\end{align}
where $acc(B_i)$ is defined as the average accuracy of the predictions in bin $B_i$.\\
\indent The bin boundaries are, in our case, defined  as equally spaced intervals, and $M=15$.\\ 
\indent Following the same argument given for \textit{isotonic regression}, \textit{histogram binning} does not affect the accuracy of the original model.

\section*{Notes on the ECE}

As stated in the main document, evaluating uncertainty calibration with the ECE can generate trivial uninformative solutions that obtain an optimal score, being one of the main criticisms about this metric in relevant literature. The following subsection presents an example of such phenomena.

\subsection*{Example}

Let us consider a pair of random variables $(X,Y)$ ($X$ represents an input space and $Y$ the corresponding set of true labels) and a model $f: X \rightarrow \Delta_3$ ($\Delta_3$ is a 2 dimensional probability simplex), such that, for a set of possible classes $\{a,b,c\}$,
\begin{align}
\mathds{P} [Y = a \ | \ f(X)] = 0.5, \quad \mathds{P} [Y = b \ | \ f(X)] = \mathds{P} [Y = c \ | \ f(X)] = 0.25
\end{align}
and
\begin{align}
    f(X) = \left(0.5, 0.3, 0.2 \right).
\end{align}
Such model gives no useful information in the task of classifying a given input, because it will always predict the same class, indiscriminately. However, given the law of large numbers, we will have that for a binning scheme $\{ B_1, B_2, \ldots, B_M \}$,
\begin{align}
\lim_{N \to \infty} ECE = \lim_{N \to \infty} \sum_{i=1}^M \frac{|B_i|}{N} | conf(B_i) - acc(B_i) | = 0.
\end{align}

\bibliographystyle{abbrv}
\bibliography{ref}